\documentclass[10pt, a4paper]{article}

\usepackage[final]{lrec2026} 

\usepackage{multirow}
\usepackage{url}
\usepackage{tcolorbox}
\usepackage{booktabs}
\usepackage{adjustbox}
\usepackage{amsmath}
\usepackage{subcaption}
\usepackage{hyperref}

\title{Do Large Language Models Grasp The Grammar? Evidence from Grammar-Book-Guided Probing in Luxembourgish}

\begingroup
\centering
\name{
\noindent\parbox{1.0\textwidth}{
\centering
\textbf{Lujun Li\textsuperscript{$\spadesuit$1}, Yewei Song\textsuperscript{$\spadesuit$1}, Lama Sleem$^{1}$,
Yiqun Wang$^{1}$, \\Yangjie Xu$^{1}$, Cedric Lothritz$^{2}$,
Niccolò Gentile$^{3}$, Radu State$^{1}$,\\
Tegawendé F. Bissyandé$^{1}$, and Jacques Klein$^{1}$
}
}
}
\endgroup

\address{
$^{1}$University of Luxembourg \\
$^{2}$Luxembourg Institute of Science and Technology \\
$^{3}$Foyer S.A.\\[3pt]
}

\abstract{
Grammar refers to the system of rules that governs the structural organization and the semantic relations among linguistic units such as sentences, phrases, and words within a given language. In natural language processing, there remains a notable scarcity of grammar-focused evaluation protocols, a gap that is even more pronounced for low-resource languages. Moreover, the extent to which large language models genuinely comprehend grammatical structure especially the mapping between syntactic structures and meanings remains under debate. To investigate this issue, we propose a Grammar-Book–Guided evaluation pipeline intended to provide a systematic and generalizable framework for grammar evaluation consisting of four key stages, and in this work we take Luxembourgish as a case study. The results show a weak positive correlation between translation performance and grammatical understanding, indicating that strong translations do not necessarily imply deep grammatical competence. Larger models perform well overall due to their semantic strength but remain weak in morphology and syntax, struggling particularly with Minimal Pair tasks, while strong reasoning ability offers a promising way to enhance their grammatical understanding. The code and detailed prompts are open-sourced at : \url{https://github.com/DobricLilujun/GBGP.git}.
 \\ \newline  }

\begin{document}

\maketitleabstract
\section{Introduction}

\def\thefootnote{$\spadesuit$}\footnotetext{These authors contributed equally to this work.}\def\thefootnote{\arabic{footnote}}


In recent years, Large Language Models (LLMs) have transformed the field of Natural Language Processing (NLP), achieving remarkable progress across a variety of tasks such as text generation and translation \cite{minaee2024large}. Their ability to produce coherent, and contextually relevant text has raised important discussions about whether such models merely mimic language patterns or demonstrate a deeper understanding of linguistic structures. Furthermore, while much of the early focus has been on their fluency and performance in high-resource languages such as English, French, or German, less attention has been paid to their grammatical competence in low-resource languages \cite{li2024quantifying}.

\noindent \textbf{Grammar Challenges} Machine translation, in particular, is an area where LLMs are widely applied and evaluated. Success is often measured in terms of fluency and semantic adequacy compared to the reference (golden) translation; however, these metrics do not necessarily capture whether the model adheres to the explicit grammatical rules of the target language\cite{caglayan2020curious}. For widely used languages with abundant training data, grammar is often implicitly learned through statistical exposure. However, the challenge becomes much more pronounced for low-resource languages, which have limited digital representation and many corner cases that are especially difficult to capture during training\cite{li2404language,zhong2024opportunities,cahyawijaya2024llms}. In such cases, an LLM’s ability to generate correct translations may not imply genuine grammatical understanding. In this sense, Luxembourgish presents a compelling case study in this context. As a low-resource language with fewer available corpora compared to its neighboring languages (German and French), Luxembourgish can pose difficult challenges on both translation and language understanding\cite{lothritz2022luxembert}. 

\noindent \textbf{Research Focus} As a result, first and foremost, a key question arises: \textbf{\textit{Do LLMs truly grasp the underlying grammar of a language, or are they simply relying on superficial translation pattern recognition that overlooks deeper syntactic structures?}} To address this question, in this work, a grammar probing pipeline is introduced to evaluate the grammatical competence of LLMs in Luxembourgish. The proposed approach goes beyond measuring translation fluency, by instead explicitly incorporating linguistic theory. To support our evaluation, we use a Luxembourgish grammar document that outlines rules such as syntactic characteristics and lexical structures. Based on this resource, we create a dataset containing grammar point descriptions, Luxembourgish example sentences, and their English translations as reference data for Luxembourgish-to-English translation\cite{Luxembourgish}. 

\noindent \textbf{Paper Overview} 
The main contribution of this work lies in the proposed evaluation pipeline and the resulting insights into LLM behavior in the grammar of low-resource languages. These elements form the foundation for a series of controlled probing experiments aimed at assessing whether LLMs align their translations with explicit grammatical rules. This approach offers a new perspective on assessing LLMs not only as fluent generators but as systems whose outputs require grammatical validation. The paper is structured as follows: Section 2 reviews related work, Section 3 presents the evaluation pipeline and the metrics, Section 4 reports results and analysis, Section 5 presents conclusions and implications, and Section 6 provides the discussion.

\section{Related Work}
\noindent \textbf{BERT Moment in Translation} LLM remain controversial regarding whether they truly understand semantics or merely mimic human speech; a potential attitude that makes one wonder whether an LLM does not amount to being just a ``stochastic parrot'' \citet{10.1145/3442188.3445922}. Despite these debates, their performance in language translation has become unparalleled, vastly exceeding many previous non-transformer and even encoder–decoder models \cite{tiedemann_democratizing_2024,nllbteam2022languageleftbehindscaling}. This development constitutes a genuine “BERT moment” for translation research. Previous non-conversational models showed limited logical reasoning, and their multilingual ability was mostly attributed to abstract mechanisms like semantic mapping. In contrast, LLMs mark a major change: by asking targeted questions and running controlled comparisons, one can directly test how they understand and apply grammatical rules, giving clearer, verifiable evidence of their cross lingual abilities.

\noindent \textbf{Grammar Lags Behind} In terms of syntactic comprehension, LLMs gradually learn hidden states which represent linguistic structures during training throughout their layers. As model size and training data increase, such syntactic differentiation becomes more pronounced—indicating that the model progressively acquires a form of syntactic understanding \cite{duan2025syntaxspecializationemergeslanguage}. Nonetheless, in multilingual contexts, LLMs are prone to ``overcorrection'' or ``hallucinations'', generating syntactically correct but semantically implausible sentences, which highlight their limitations in reasoning and ambiguity resolution \cite{10.1371/journal.pone.0312881}. This suggests occasional failures or neglect in their grasp of syntax. For low-resource languages and dialects, LLMs' performances on syntactic tasks remains relatively weak, calling for greater data diversity and enrichment \cite{zhang-etal-2025-milic}. Several open-source datasets provide linguistic minimal pairs (MPs: \cite{warstadt-etal-2020-blimp-benchmark, xiang-etal-2021-climp}) and leverage them to evaluate the core grammatical competence of LLMs \cite{ide-etal-2025-make}. Nevertheless, to the best of current knowledge, a unified framework for systematically and comprehensively assessing LLMs’ syntactic understanding has not yet been established.





\section{Method}

\subsection{Mindsets - Why Grammar Book?}

\begin{figure*}[!htbp]
    \centering
    \includegraphics[width=1.00\textwidth]{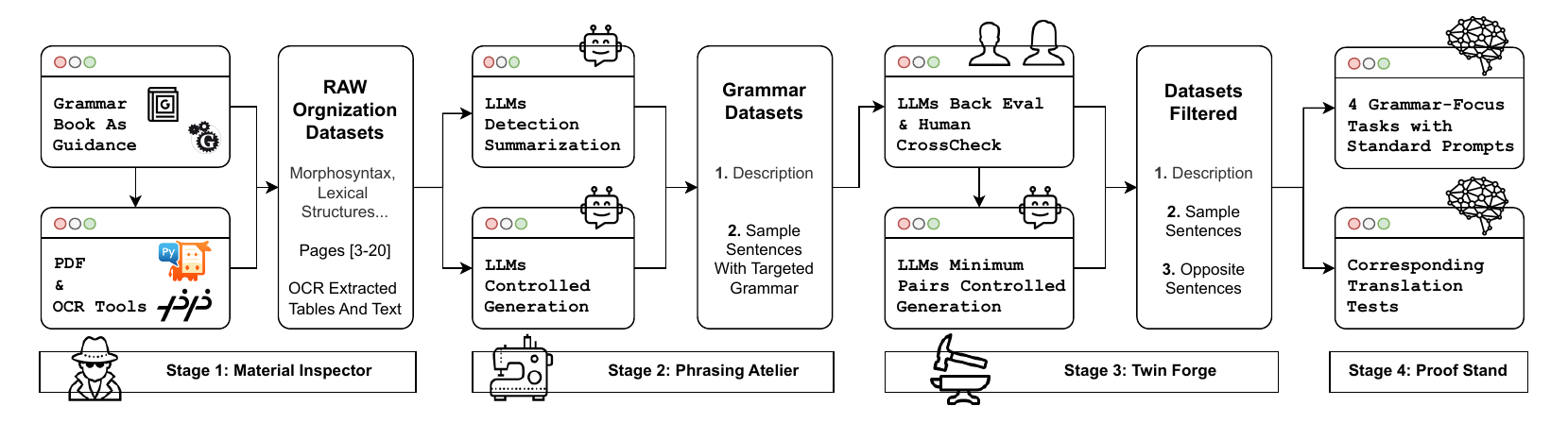}
    \caption{The proposed framework}
    \label{fig:pipeline}
\end{figure*}

\noindent \textbf{Translation enhanced by Grammar} Our work is inspired by "Machine Translation from One Book" which demonstrates that it is possible to translate a language with virtually no web data (Kalamang) using a grammar book as the primary resource \cite{Tanzer2023ABF,kondo-etal-2024-enhancing}. However, \cite{Aycock2024CanLR} shows that the translation performance in this setting is primarily attributable to the parallel example sentences embedded within the grammar book, rather than the grammatical descriptions themselves. This raises the question: instead of using the grammar book for translation, can we leverage a complete grammar book to simply evaluate the grammatical understanding capabilities of LLMs? Furthermore, open-source, grammar-focused datasets remain scarce and often exhibit incomplete coverage of grammatical phenomena, with many evaluations restricted to a single minimal-pair testing paradigm whose richness is limited.

\noindent \textbf{Grammar coverage is comprehensive} Additionally, for every actively used language, grammar books possesses also lexicon sections, and grammar references typically enumerate the vast majority of grammar points alongside representative examples, yielding informational breadth sufficient. This motivates adopting grammar books as the foundation for assessing models’ grammatical understanding, thus explains the choice of leveraging said books as the primary resource.

\subsection{Pipeline Structure}

We introduce a grammar-grounded, end-to-end evaluation pipeline for LLMs that begins from a canonical grammar reference and systematically probes grammar-related competencies across multiple steps, as shown in Figure \ref{fig:pipeline}.

\begin{enumerate}
    \item \textbf{Material Inspector}: component which identifies and typifies grammatical phenomena and error patterns from the grammar document;
    \item \textbf{Phrasing Atelier}: component which distills relevant grammatical content, extracts targeted constructions such as grammar description, and elicits constrained generations to specific forms as language pairs;
    \item \textbf{Twin Forge}: component which conducts back checking and filtering from step 2 and constructs minimal pairs for grammar understanding evaluations;
    \item \textbf{Proof Stand}: component which operationalizes grammatical competence in end-to-end tasks for comprehensive grammar understanding evaluation of LLMs.
\end{enumerate}
   
\noindent\textbf{Material Inspector}: Primarily, this step extracts data from a grammar book in PDF format, distinguishing between image-only PDFs and text-recognizable PDFs. For image-only files, PaddleOCR \cite{cui2025paddleocr30technicalreport} is employed to perform page-level image extraction, followed by table detection and optical character recognition. For text-recognizable files, PyMuPDF\cite{PyMuPDF} is used to identify chapters based on font-size heuristics and to extract text on a per-chapter basis; tables in this case are still processed to localize them and recover both structure and content. In extracting information from grammar books, while grammar descriptions are essential, substantial information is often hidden in or conveyed through numerous tables, so the pipeline places particular emphasis on robust table detection and recognition. 

\noindent\textbf{Phrasing Atelier} This stage detects and synthesizes grammatical points by leveraging the spatial distribution of descriptions and tables. First, sections of the grammar book that cover syntax and morphology are filtered, given their relevance to translation and precise expression. The text is then segmented using sentence-level sliding windows to minimize omissions of grammatical points during extraction. For tables, each row is traversed and accompanied by a fixed-length window of surrounding textual context supplied to LLMs, which analyze the content row by row to identify and summarize embedded grammatical points. Additionally, based on the extracted grammatical points, the system prompts the model to design pairs of English-to-Luxembourgish sentences of controlled length that obligatorily instantiate the identified grammatical targets.

\noindent\textbf{Twin Forge} In this phase, the extracted grammatical targets undergo back testing: LLMs generated example sentences from step 2 will be scored via translation quality and assessed for whether they instantiate the intended grammatical rule; items receiving a negative judgment are discarded. Moreover, for each surviving grammatical point and its example sentence, the system elicits a corresponding ungrammatical item that contrast specifically in grammatical structure, enabling further fine-grained validation and instruction.

\noindent\textbf{Proof Stand} We design four comprehensive experiments to assess whether the model can understand grammatical phenomena and to delineate the boundaries of its grammatical competence.\par

\noindent\textbf{Task 1: Sentence-level multi-grammar}. Given a sentence \( s \) and a set of different grammar point descriptions \( G = \{g_1, g_2, \dots, g_n\} \), with size denoted as \( N_{1g} \), the task is to identify the subset \( G_s \subseteq G \) instantiated in \( s \). Formally, \( f_1: (s, G) \mapsto G_s \).

\begin{tcolorbox}[
colback=gray!5!white,
colframe=gray!80!black,
title=Task 1 Example Prompt Simplified,
fonttitle=\bfseries,
fontupper=\small, 
arc=2mm,
boxrule=0.3pt
]
Carefully examine the Luxembourgish sentence and its English equivalent.\\
Identify which grammar description best matches the structure or expression used in the Luxembourgish sentence\
\end{tcolorbox}

\noindent\textbf{Task 2: Grammar-level multi-sentence} Given a grammatical category \( g \) and a collection of sentences \( S = \{s_1, s_2, \dots, s_m\} \) with size denoted as \( N_{2s} \), the task is to identify the subset \( S_g \subseteq S \) conforming to \( g \). Formally, \( f_2: (g, S) \mapsto S_g \).

\begin{tcolorbox}[
colback=gray!5!white,
colframe=gray!80!black,
title=Task 2 Example Prompt Simplified,
fonttitle=\bfseries,
fontupper=\small, 
arc=2mm,
boxrule=0.3pt
]
Carefully examine each Luxembourgish sentence.\\
Identify which sentences demonstrate the given grammar knowledge point.
\end{tcolorbox}

\noindent\textbf{Task 3: Multi-sentence multi-grammar} Given a set of sentences \( S = \{s_1, \dots, s_m\} \) with size denoted as \( N_{3s} \) and grammatical categories \( G = \{g_1, \dots, g_n\} \) with size denoted as \( N_{3g} \), the task is to infer the subset \( G_S \subseteq G \) instantiated across \( S \). Formally, \( f_3: (S, G) \mapsto G_S \).

\begin{tcolorbox}[
colback=gray!5!white,
colframe=gray!80!black,
title=Task 3 Example Prompt Simplified,
fonttitle=\bfseries,
fontupper=\small, 
arc=2mm,
boxrule=0.3pt
]
Carefully examine the Luxembourgish paragraph.\\
Identify 2 grammar points from the provided list are demonstrated in this paragraph.\\
\{\{Provided grammar point list\}\}
\end{tcolorbox}

\noindent\textbf{Task 4: Identification from minimal pairs.}
Given a minimal pair \((s_c, s_i)\), where \(s_c\) denotes the correct sentence and \(s_i\) the incorrect sentence, the task is to identify the grammatical acceptable sentence. Formally, define \(f_4: (s_c, s_i) \mapsto s_c\).

\begin{tcolorbox}[
colback=gray!5!white,
colframe=gray!80!black,
title=Task 4 Example Prompt Simplified,
fonttitle=\bfseries,
fontupper=\small, 
arc=2mm,
boxrule=0.3pt
]
Carefully examine these two Luxembourgish sentences.\\
Identify which Luxembourgish sentence is grammatically acceptable.
\end{tcolorbox}







\subsection{Models and Datasets}

For document extraction, the OCR pipeline primarily utilizes a layout detection model, \textit{PP-DocLayout-L}, a table classification network, \textit{PP-LCNet\_x1\_0\_table\_cls}, and a text recognition module designed for Latin scripts, denoted as \textit{text\_recognition\_model\_name = ``latin\_PP-OCRv5\_mobile\_rec''}. For controlled retrieval and generation tasks that target specific grammatical phenomena, \textit{GPT-5} is consistently employed. Carefully engineered prompts are introduced to support information extraction and grammar-constrained sentence generation.

Following the grammatical extraction process, a dataset comprising grammatical descriptions and three to four translation pairs (from English to Luxembourgish, each annotated with relevant grammatical features) was constructed. The dataset encompasses 673 distinct grammar points and includes a total of 2,040 example sentences. Regarding the back-checking process conducted using \textit{GPT-5}, 93.97\% of the sentences were identified as grammatically correct. Furthermore, the average translation performance of the Luxembourgish-to-English sentence pairs, evaluated under the \textit{LLM-as-a-Judge} framework on a scale from 0 to 10, achieved a score of 7.9.

\begin{tcolorbox}[
colback=gray!5!white,
colframe=gray!80!black,
title=Luxembourgish Sentence,
fonttitle=\bfseries,
fontupper=\small, 
arc=2mm,
boxrule=0.3pt
]
De klengen Hond, dee nieft eisem Gaart wunnt, ass ëmmer frëndlech mat de Leit an huet d'Gewunnecht säi Schwanz héich ze drécken wann e neue Visiteur kënnt.
\end{tcolorbox}

\begin{tcolorbox}[
colback=gray!5!white,
colframe=gray!80!black,
title=English Sentence,
fonttitle=\bfseries,
fontupper=\small, 
arc=2mm,
boxrule=0.3pt
]
The small dog that lives next to our garden is always friendly with people and has the habit of wagging its tail high whenever a new visitor arrives at the house.
\end{tcolorbox}

\begin{tcolorbox}[
colback=gray!5!white,
colframe=gray!80!black,
title=Extracted Grammar Descriptions,
fonttitle=\bfseries,
fontupper=\small, 
arc=2mm,
boxrule=0.3pt
]
Because nouns themselves no longer show case, definite and indefinite articles and attributive adjectives change to mark syntactic roles. For masculine singular, articles typically alternate (e.g., \textit{de/den/dem}) to indicate nominative, accusative and dative respectively; adjectives within the phrase agree and help signal the function of the phrase in the clause. This system means word order and article/adjective morphology are crucial for identifying subjects, objects and indirect objects.
\end{tcolorbox}

As a result, we also constructed a unified translation dataset, called MIXED, as shown in Table \ref{tab:datasets}. MIXED includes 2,040 example sentences from grammar book \textbf{(G2040)}, extracted according to the grammar book, 1,012 example sentences from \textbf{FLORES200} with English and Luxembourgish pairs, and 300 long sentences \textbf{(LONG300)}, collected from open-source data, on which the model was tested to obtain comparative results and an evaluation of the model’s translation performance. The final dataset includes, as such, 3,352 pairs. 

The Luxembourgish-to-English pairs in MIXED derived from grammar books were primarily generated by large language models (LLMs) during the extraction process, which simultaneously produced the corresponding English translations. In contrast, FLORES200 contains 1,012 relatively short sentence pairs that were manually verified. The LONG300 subset includes long Luxembourgish sentences sampled from the web, translated into English using LLMs, and subsequently verified by human reviewers.

\begin{table}[h!]
\centering
\begin{adjustbox}{width=0.5\textwidth,center}
\begin{tabular}{ccccl}
\toprule
\textbf{Dataset} & \textbf{Avg. Char Len ( ± $\sigma$)} & \textbf{\# N} & \textbf{Topics} & \textbf{Ref.} \\
\midrule
\textbf{G2040}     & 171.63 ± 21.42 & 2040 & Grammars & \citeyear{Luxembourgish} \\
\textbf{LONG300}   & 429.47 ± 49.18 & 300  & News     & \citeyear{rtllu2025} \\
\textbf{FLORES200} & 125.61 ± 41.18 & 1012 & Diverse  & \citeyear{nllbteam2022languageleftbehindscaling} \\
\bottomrule
\end{tabular}
\end{adjustbox}
\caption{Summary of datasets, showing average length, entry count, and topic coverage, named \textbf{MIXED}.}
\label{tab:datasets}
\end{table}

We also selected a wide range of state-of-the-art (SOTA) open-source and closed-source models to conduct comprehensive experiments and minimize bias. Specifically, we chose three groups of models: the Gemma series, the Llama series, and the Qwen series, as shown in Table \ref{tab:models}.

\begin{table}[h!]
\centering
\begin{adjustbox}{width=0.5\textwidth,center}
\begin{tabular}{lcc}
\toprule
Model & Model Size & Release Date \\
\midrule
GPT-5 & $\sim$300B & 2025-08 \\
google/gemma-2-2b-it & 2B & 2024-06 \\
google/gemma-2-9b-it & 9B & 2024-06 \\
google/gemma-3-4b-it & 4B & 2025-03 \\
google/gemma-3-12b-it & 12B & 2025-03 \\
google/gemma-3-27b-it & 27B & 2025-03 \\
meta-llama/Llama-3.2-1B-Instruct & 1B & 2024-04 \\
meta-llama/Llama-3.2-3B-Instruct & 3B & 2024-04 \\
meta-llama/Llama-3.3-70B-Instruct & 70B & 2024-10 \\
Qwen/Qwen3-4B-Instruct-2507 & 4B & 2025-07 \\
Qwen/Qwen3-30B-A3B-Instruct-2507 & 30B & 2025-07 \\
Qwen/Qwen3-Next-80B-A3B-Instruct & 80B & 2025-09 \\
Qwen/Qwen3-Next-80B-A3B-Thinking & 80B & 2025-09 \\
\bottomrule
\end{tabular}
\end{adjustbox}
\caption{Major open-source models selected}
\label{tab:models}
\end{table}

\subsection{Metrics}

\textbf{CometScore} is a neural network–based evaluation metric , designed to assess machine translation quality by jointly modeling fluency, adequacy, and semantic preservation \cite{rei-etal-2020-comet}. It provides both reference-based and reference-free variants, aiming to approximate human judgment more faithfully than surface-level lexical overlap measures. Owing to the current lack of support for Luxembourgish in the reference-free version, this study employs the reference-based configuration of CometScore, specifically the \texttt{Unbabel/wmt23-cometkiwi-da-xl} model.

\textbf{SentencePiece BLEU (spBLEU)} is a classical machine translation evaluation metric that measures the degree of word or phrase (\(n\)-gram) overlap between the machine translation output and the reference translation \cite{kudo-richardson-2018-sentencepiece}. It calculates precision and integrates a length penalty.

\textbf{ChrF++} is based on the F-score computed over character-level (rather than word-level) \(n\)-grams, combining both character and word \(n\)-gram matching \cite{popovic-2017-chrf} allowing it to more flexibly reflect translation similarity.

\textbf{For Tasks 1 through 4}, we used accuracy as the primary metric to evaluate the model's grammatical understanding, supplemented by standard deviation (std). We further examined how the model's performance in selecting the correct option varied with different candidate set sizes.

\section{Results}

\subsection{Pipeline Validity Checking}

We randomly selected 50-100 sentences in both English and Luxembourgish and sent them to local volunteers to verify whether the Luxembourgish sentences, generated based on specific grammar points, faithfully reflected the intended grammatical structures. Our results show that 98\% of the sentences contained the corresponding grammar. Notably, when designing grammar points where pronunciation directly influences spelling or word formation - such as the Luxembourgish ``Eifeler Regel''- we encountered additional complexity. The ``Eifeler Regel'' is a phonological and orthographic rule involving the omission or retention of a terminal *n* in certain words depending on the initial letter of the following word, with spelling adapted to match actual pronunciation. This issue is frequently discussed, and within our generated sentences, violations of the ``Eifeler-Regel” accounted for 32\% \cite{inbook}. Regardless, we still marked these as grammatically correct, as this phenomenon was outside the grammar points under our focus. This finding highlights that rules derived from pronunciation are especially difficult for LLMs to learn, and display when they lack phonological awareness. Furthermore, it reflects the broader challenge for LLMs to acquire complete and rigorous grammatical rules in low-resource languages.

\begin{table*}[!htbp]
  \centering
  \begin{adjustbox}{max width=\textwidth}
  \begin{tabular}{lccccccc}
    \toprule
    \multirow{2}{*}{\textbf{Selected Model}} &
      \textbf{\begin{tabular}[c]{@{}c@{}}CometScore \\ ($\mu$ $\pm$ std)\end{tabular}} &
      \textbf{\begin{tabular}[c]{@{}c@{}}spBLEU\\ ($\mu$ $\pm$ std)\end{tabular}} &
      \textbf{\begin{tabular}[c]{@{}c@{}}ChrF++\\ ($\mu$ $\pm$ std)\end{tabular}} &
      \textbf{\begin{tabular}[c]{@{}c@{}}Task 1 Acc\\ ($\mu$ $\pm$ std)\end{tabular}} &
      \textbf{\begin{tabular}[c]{@{}c@{}}Task 2 Acc\\ ($\mu$ $\pm$ std)\end{tabular}} &
      \textbf{\begin{tabular}[c]{@{}c@{}}Task 3 Acc\\ ($\mu$ $\pm$ std)\end{tabular}} &
      \textbf{\begin{tabular}[c]{@{}c@{}}Task 4 Acc\\ ($\mu$ $\pm$ std)\end{tabular}} \\
    &
      \textbf{\begin{tabular}[c]{@{}c@{}}MIXED\\ \#3000\end{tabular}} &
      \textbf{\begin{tabular}[c]{@{}c@{}}MIXED\\ \#3000\end{tabular}} &
      \textbf{\begin{tabular}[c]{@{}c@{}}MIXED\\ \#3000\end{tabular}} &
      \begin{tabular}[c]{@{}c@{}}Base\\ Settings\end{tabular} &
      \begin{tabular}[c]{@{}c@{}}Base\\ Settings\end{tabular} &
      \begin{tabular}[c]{@{}c@{}}Base\\ Settings\end{tabular} &
      \begin{tabular}[c]{@{}c@{}}Base\\ Settings\end{tabular} \\
    \midrule
        GPT-5 &
      0.26 ± 0.13 &
      \textbf{39.30 ± 14.46} &
      \textbf{56.20 ± 11.01} &
      0.94 ± 0.01 &
      \textbf{0.95 ± 0.01} &
      \textbf{0.91 ± 0.01} &
      \textbf{0.61 ± 0.02}  \\
    \midrule
    google/gemma-2-2b-it &
  0.04 ± 0.11 &
  3.07 ± 3.82 &
  17.80 ± 5.67 &
  0.80 ± 0.01 &
  0.87 ± 0.01 &
  0.60 ± 0.02 &
  0.50 ± 0.01 \\
    google/gemma-2-9b-it &
      \underline{0.18 ± 0.13} &
      \underline{11.28 ± 8.48} &
      \underline{31.87 ± 8.81} &
      \underline{0.83 ± 0.01} &
      \underline{0.92 ± 0.01} &
      \underline{0.83 ± 0.01} &
      \underline{0.51 ± 0.01} \\
    \midrule
       google/gemma-3-4b-it &
      0.14 ± 0.11 &
      10.09 ± 7.50 &
      31.07 ± 7.63 &
      0.81 ± 0.01 &
      0.79 ± 0.02 &
      0.47 ± 0.02 &
      0.52 ± 0.01 \\
        google/gemma-3-12b-it &
      \underline{0.22 ± 0.12} &
      \underline{23.06 ± 11.83} &
      \underline{44.17 ± 9.77} &
      0.87 ± 0.01 &
      \textbf{\underline{0.95 ± 0.01}} &
      0.85 ± 0.01 &
      \underline{0.54 ± 0.01} \\
        google/gemma-3-27b-it &
      0.22 ± 0.13 &
      17.90 ± 10.56 &
      39.75 ± 9.60 &
      \underline{0.88 ± 0.01} &
      \textbf{\underline{0.95 ± 0.01}} &
      \underline{0.87 ± 0.01} &
      \underline{0.54 ± 0.01} \\
    \midrule
    meta-llama/Llama-3.2-1B-Instruct &
  0.08 ± 0.15 &
  1.72 ± 2.79 &
  13.16 ± 5.83 &
  \underline{0.88 ± 0.01} &
  0.59 ± 0.02 &
  0.43 ± 0.02 &
  \underline{0.51 ± 0.01}  \\
    meta-llama/Llama-3.2-3B-Instruct &
  0.06 ± 0.10 &
  3.79 ± 4.42 &
  21.30 ± 6.28 &
  0.84 ± 0.01 &
  0.75 ± 0.02 &
  0.62 ± 0.02 &
  0.47 ± 0.01  \\
    meta-llama/Llama-3.3-70B-Instruct &
  \underline{0.22 ± 0.12} &
  \underline{26.52 ± 12.84} &
  \underline{46.87 ± 10.34} &
  \underline{0.88 ± 0.01} &
  \underline{0.95 ± 0.01} &
  \underline{0.84 ± 0.01} &
  \underline{0.51 ± 0.01} \\
    \midrule
    Qwen/Qwen3-4B-Instruct-2507 &
  0.15 ± 0.13 &
  8.12 ± 7.04 &
  26.74 ± 7.55 &
  0.86 ± 0.01 &
  0.90 ± 0.01 &
  0.82 ± 0.01 &
  0.52 ± 0.01 \\
    Qwen/Qwen3-30B-A3B-Instruct-2507 &
  0.26 ± 0.13 &
  14.40 ± 10.05 &
  35.84 ± 8.85 &
  0.85 ± 0.01 &
  \textbf{\underline{0.95 ± 0.01}} &
  0.81 ± 0.01 &
  0.54 ± 0.01\\
  Qwen/Qwen3-Next-80B-A3B-Instruct &
  \underline{\textbf{0.26 ± 0.12}} &
  \underline{19.15 ± 11.46} &
  40.31 ± 9.59 &
  0.92 ± 0.01 &
  \textbf{\underline{0.95 ± 0.01}} &
  \underline{0.88 ± 0.01} &
  \underline{0.57 ± 0.01} \\
Qwen/Qwen3-Next-80B-A3B-Thinking &
  0.24 ± 0.09 &
  18.53 ± 10.60 &
  \underline{40.41 ± 8.70} &
  \textbf{\underline{0.95 ± 0.01}} &
  \textbf{\underline{0.95 ± 0.01}} &
  \textbf{\underline{0.91 ± 0.01}} &
  \underline{0.58 ± 0.01} \\
    \bottomrule
  \end{tabular}
  \end{adjustbox}
  \caption{Performance comparison of selected models on the MIXED benchmark, reporting translation metrics from English to Luxembourgish(CometScore, spBLEU, ChrF++) and downstream task accuracies ($\mu$ ± std) under base settings.}
  \label{tab:main_perf}
\end{table*}

\subsection{Statistical Analysis}

\begin{figure}[htbp]
    \centering
    \includegraphics[width=0.51\textwidth]{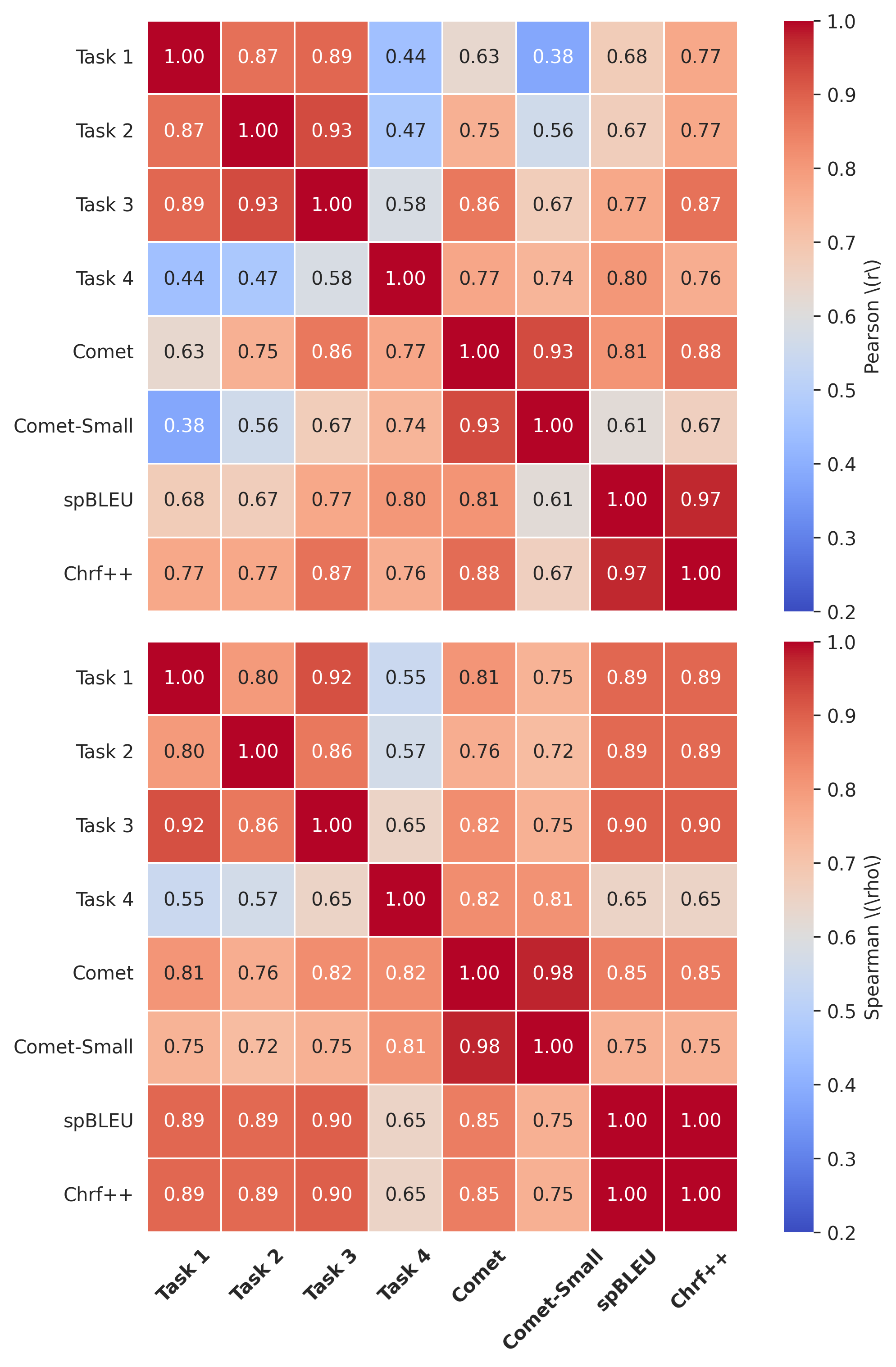} 
    \caption{Side-by-side Pearson (top) and Spearman (bottom) correlations among grammar understanding and translation metrics; values are annotated and mapped to a centered diverging color map.}
    \label{fig:heatmap}
\end{figure}

We conducted Spearman and Pearson correlation analyses on the performance of the four grammatical comprehension test tasks as well as on translation performance, and the results are illustrated in Figure \ref{fig:heatmap}. It can be observed that Tasks 1, 2, and 3 exhibit very strong correlations, primarily reflecting the ability to identify grammatical structures and their corresponding sentences. However, these three tasks show relatively low correlations with Task 4, which involves identifying sentences with correct grammatical usage. As a result, there is a weak positive correlation between performance on grammatical comprehension tasks and translation performance; however, it is insufficient to conclude that strong translation implies grammatical understanding.

In addition, we included the relatively smaller \texttt{Unbabel/wmt22-cometkiwi-da} (Small size) model as a supplementary evaluation to the \texttt{Unbabel/wmt23-cometkiwi-da-xl} (Big size) model. We found that the larger model shows higher correlations with spBLEU and ChrF++, while the smaller model’s evaluation scores are less stable. This finding led us to adopt \texttt{Unbabel/wmt22-cometkiwi-da} as our primary evaluation metric.

\noindent \textbf{Weak Correlated} This correlation analysis indicates that a model’s grammatical understanding and its translation performance are only weakly correlated. A model with strong translation performance does not necessarily have strong grammatical comprehension. More in general, we provide a statistical experiment–based analysis verifying that machine translation tends to resemble linguistic imitation rather than genuine understanding. Furthermore, we also observed that the performance trend of the minimal pair grammatical tests differs from that of other performance evaluations, suggesting that it merits deeper investigation. Finally, the smaller cometkiwi model exhibits greater instability in translation results. Thus, we recommend using comparatively larger models to achieve more consistent and reliable evaluation results.

\subsection{Translation Vs Grammar Understanding?}

The central question we aim to clarify is whether translation is equivalent to deep grammatical understanding. Our prior assumption was that if a model translates well, it should, to some extent, demonstrate an understanding of grammar, since training on large-scale parallel corpora implicitly exposes the model to numerous grammatical patterns. Consequently, the model, through its internal reasoning and comprehension mechanisms, might inherently strengthen its grammatical competence. 

In Table \ref{tab:main_perf}, the experiments are primarily conducted under the base setting. In Task 1, we set \( N_{1g} = 2 \); in Task 2, \( N_{2s} = 1 \); in Task 3, \( N_{3g} = 4 \) and \( N_{3s} = 2 \); and in Task 4, we use pairs of sentences with correct and incorrect grammatical usage, along with the corresponding grammar descriptions that need to be verified.

\noindent \textbf{Partial mastery of grammar} We take the Llama family as a case study and observe that scaling from 1B - 3B to 70B parameters improves translation performance from spBLEU $1.72$ to $26.52$, which yields partial gains on Task 2 and Task 3 (approximately 30\%), yet leaves Task 1 essentially unchanged and Task 4 still near chance-level performance. Comparable patterns emerge for the Qwen and Gemma families. Only GPT-5 achieves consistently superior results across all tasks and surpasses chance performance on Task 4. In our evaluation of Task 4, we prompted LLMs to directly generate the grammatically correct Luxembourgish sentence; however, given that grammar constitutes a strongly logical rule system, we posit that incorporating Chain-of-Thought (CoT) reasoning may enhance model performance on this task, whereas direct inference to the final option could lead the model to select responses intuitively rather than through rigorous logical deduction. Moreover, interestingly, reasoning model ``Qwen3-Next-80B-A3B-Thinking'' demonstrate higher performance on most grammar understanding tasks, even outperforming GPT-5, highlighting the crucial role of logical reasoning in grammatical comprehension. Moreover, increasing model size within the same family confers modest benefits to grammatical understanding - likely via enhanced reasoning - yet these gains are markedly smaller than those observed for translation.

\subsection{Model Point of Views?}

\begin{figure*}[!htbp] 
    \centering
    \includegraphics[width=\textwidth]{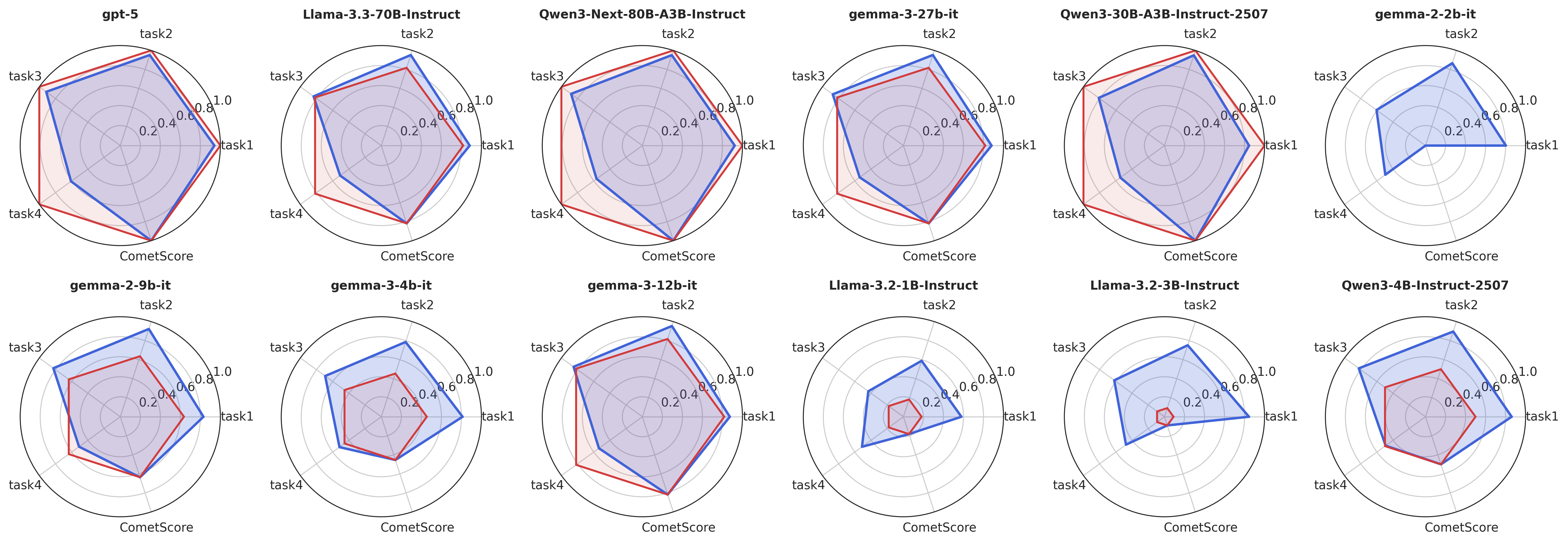} 
    \caption{A radar chart comparing 12 instruction-tuned LLMs across four task dimensions. Concentric rings represent normalized model-specific CometScores (in red), serving as an external baseline for translation fluency. The red ring visually illustrates the relative performance gap from translation to Task 1 through Task 4. All metrics are normalized to a 0–1 scale.}
    \label{fig:radar_plot}
\end{figure*}

We visualize the models individually in Figure \ref{fig:radar_plot} and observe that on Task 4, located in the lower-left quadrant, models consistently performs poorly, further underscoring a deficit in distinguishing correct from incorrect sentences and a mismatch between translation performance and grammatical understanding. For instance, Qwen-3-Next-80B display similar with GPT-5 in both translation and grammatical competence. Overall, when models are very small, translation performance can lag behind grammatical understanding. However, as model size increases, translation capabilities strengthen and can possibly surpass grammatical understanding. Moreover, translation improvements appear to encounter capacity-related bottlenecks beyond a certain scale, at which point both grammatical understanding and translation plateau, as exemplified by transitioning from  Gemma-3-12B to 27B, or Qwen3-30B to larger variants.

\begin{figure*}[!htbp]
    \centering
    \begin{subfigure}{0.95\textwidth}
        \centering
        \includegraphics[width=\textwidth]{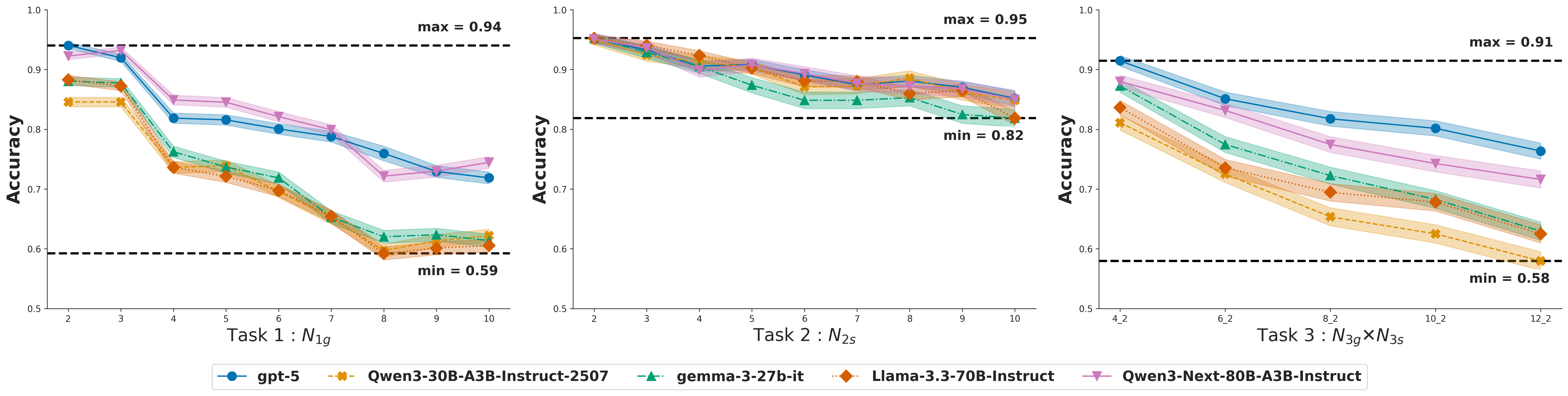}
        \caption{Big Size Models}
        \label{fig:sub1}
    \end{subfigure}
    
    \vspace*{-0.2cm} 
    
    \begin{subfigure}{0.95\textwidth}
        \centering
        \includegraphics[width=\textwidth]{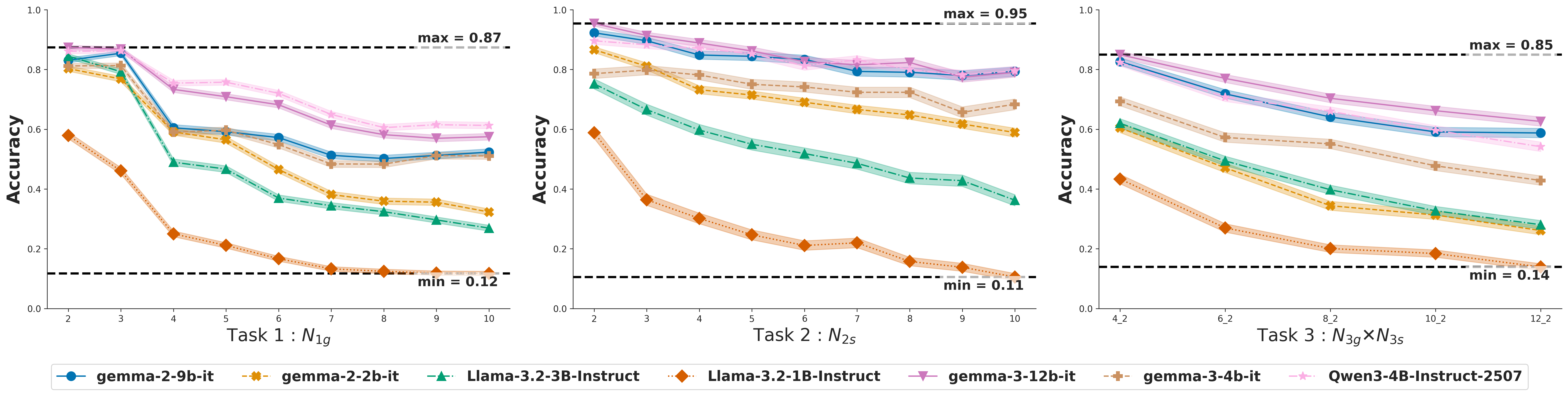}
        \caption{Small Size Models}
        \label{fig:sub2}
    \end{subfigure}
    
    \caption{The degradation in model performance as a function of the number of options presented in the prompts across different architectures.}
    \label{fig:number_choices}
\end{figure*}

\subsection{More choices, more confusion?}

We also investigate whether model choices vary with the number of options provided; selecting between two grammatical targets is naturally simpler than selecting among four, which yields the results in Figure \ref{fig:number_choices}, where several patterns are particularly noteworthy. Task 1 is markedly more sensitive than Task 2, indicating that identifying the appropriate grammar is harder for models than selecting the corresponding sentence. Sentences are comparatively easier for LLMs to interpret, whereas logically oriented, abstract grammatical categories are harder to discriminate. For larger models, Task 3 exhibits a gradual, steady decline as number of options increase, yet overall performance remains high.

Surprisingly, in Task 2 - selecting the sentence corresponding to a given grammatical target - even small models display substantial robustness. Consequently, models do demonstrate a degree of grammatical understanding, especially given that with ten grammatical targets or sentences, accuracy remains well above chance, around 80\%. Larger models show this trend more clearly: despite lacking explicit grammar-focused post training, their grammatical competence is considerable after extensive multilingual training. This is consistent with human language acquisition: extensive reading supports word formation, sentence construction, and the inductive abstraction of grammatical rules, mirroring gains in grammatical competence seen in larger LLMs. Although overlap with previously seen grammar materials cannot be ruled out, our task requires complex grammar identification and understanding; moreover, the low-resource setting substantially limits data contamination, strengthening the credibility of the results.


\section{Conclusion}

We propose a complete pipeline for assessing LLMs’ grammatical understanding, constructed as a controlled generation framework grounded in grammar textbooks and readily extensible to other languages, finally ebvaluated considering a translation task. The pipeline comprises four components - “Material Inspector,” “Phrasing Atelier,” “Twin Forge,” and “Proof Stand” - which leverage OCR and LLMs to (i) extract and normalize textbook content, (ii) synthesize controlled grammatical targets, (iii) generate corresponding example sentences, and (iv) conduct final verification. We apply this pipeline to a Luxembourgish grammar textbook to extract grammatical points and produce example sentences, yielding a range of noteworthy findings about grammatical structure and usage, considering translation as reference use case.

LLMs appear to acquire language skills in a human-like manner: extensive exposure to text during pre-training functions as “reading,” while subsequent error-corrective procedures during instruction fine-tuning play the role of “classroom” refinement, together fostering emergent grammatical competence. Empirically, a weak positive correlation between translation performance and grammatical understanding supports this view, though gains in translation do not necessarily translate into substantial improvements in grammar. As model size increases, advances in translation - i.e., sentence production - tend to outpace advances in grammatical understanding, suggesting that reading-like exposure yields only shallow grammatical knowledge and that deeper grammatical mastery requires targeted training. Finally, when grammatical tasks become more challenging, larger LLMs can still identify correct answers, whereas smaller models suffer sharp performance declines, indicating that grammar, as a logic-intensive domain, demands sufficient model capacity and reasoning ability to handle its abstract nature.

To answer the research question - \textit{Do LLMs truly grasp the underlying grammar of a language, or are they simply relying on superficial translation strategies that overlook deeper syntactic structures?} - the evidence suggests that models possess a limited degree of grammatical understanding, acquired in a human-like manner through extensive reading during pre-training. However, achieving deep grammatical competence appears to require strong logical reasoning abilities and targeted training, which current models typically lack. Consequently, their behavior is dominated by semantic-level sentence production and imitation rather than rigorous, character-level grammatical verification.

\section{Future work}

It is worth noting that Task 4 - the Minimal Pair test - remains highly challenging for models, because incorrect grammatical usage often yields sentences that are semantically similar yet lexically distinct, a predicament that cannot be resolved by semantic cues alone and instead requires finer-grained, character-level analysis and understanding. This observation motivates exploring Chain-of-Thought as a lightweight means of strengthening logical reasoning in LLMs, as well as adopting dedicated reasoning models (e.g., DeepSeek) to assess whether substantially higher performance in grammatical understanding can be achieved: pursuing these directions constitutes a promising avenue for future research. Finally, conducting grammatical evaluations across multiple languages will form another direction for future work, enabling the establishment of a benchmark for models’ grammatical understanding, especially on low-resource languages.

\section*{Acknowledgements}

All models and resources developed in this work are strictly intended for research and educational purposes according to OpenAI usage guidelines; no model weights or derivatives are used - or will be used - for any commercial application. We exclusively utilize publicly available corpora or datasets for which explicit authorization has been obtained from the original data providers.  All license terms have been reviewed to ensure full compliance with copyright, attribution, and sharing requirements.

No personally identifiable information (PII) is collected during this research. All data processing, storage, and retention policies are fully aligned with the EU General Data Protection Regulation (GDPR). All code, models, and processed data artifacts will be released under an open-source, research-oriented license (e.g., CC BY-NC), accompanied by comprehensive documentation and bias-analysis methodology to promote transparency and reproducibility.  We commit to ongoing ethical oversight through periodic reevaluation of datasets and model outputs, prompt updates in response to emerging concerns, and consultation with interdisciplinary advisory boards to ensure adherence to the highest ethical standards.

\bibliographystyle{lrec2026-natbib}
\bibliography{lrec2026-example}

\bibliographystylelanguageresource{lrec2026-natbib}
\bibliographylanguageresource{languageresource}

\end{document}